\documentclass{article}

\usepackage[preprint]{neurips_2021}

\usepackage[utf8]{inputenc} 
\usepackage[T1]{fontenc}    
\usepackage{hyperref}       
\usepackage{url}            
\usepackage{booktabs}       
\usepackage{amsfonts}       
\usepackage{nicefrac}       
\usepackage{microtype}      
\usepackage{xcolor}         
\usepackage{pgfplots}
\usepackage{amsmath}
\hypersetup{
    colorlinks=true,
    linkcolor=blue,
    citecolor=black,
    urlcolor=blue,
}

\usepackage{algorithm,algpseudocode}

\usepackage{caption}
\usepackage{subcaption}

\usepackage[frozencache,cachedir=.]{minted}

\title{Graph Neural Networks for Image Classification and Reinforcement Learning using Graph representations}

\author{%
  Naman Goyal* \\
  Department of Computer Science\\
  Columbia University\\
  New York, NY 10027 \\
  \texttt{ng2848@columbia.edu} \\
  \And
  David Steiner* \\
  Department of Computer Science\\
  Columbia University\\
  New York, NY 10027 \\
  \texttt{ds3816@columbia.edu}
}
\pgfplotsset{compat=1.17}
\begin{document}

\maketitle

\begin{abstract}
  In this paper, we will evaluate the performance of graph neural networks in two distinct domains: computer vision and reinforcement learning. In the computer vision section, we seek to learn whether a novel non-redundant representation for images as graphs can improve performance over trivial pixel to node mapping on a graph-level prediction graph, specifically image classification. For the reinforcement learning section, we seek to learn if explicitly modeling solving a Rubik's cube as a graph problem can improve performance over a standard model-free technique with no inductive bias.
\end{abstract}

\section{Introduction}
Many important real-world problems can be modeled as graphs, but due to the heterogeneous nature of graphs, which can have wildly different numbers of vertices and edges, as well as varying levels of sparsity, they are hard to encode directly to feed to a traditional feedforward network and define a gradient based optimisation mechanism for them. In this work, we aim to apply graph neural network techniques to two distinct problem domains: (1) Image classification (2) Solving a Rubik’s cube. The work is a mix of both understanding and analysis as well as exploratory research.

For image classification, the goal is to come up with an novel representation for images as graphs. The current techniques are much more ad-hoc and don't generalize well, and suffer from curse of dimensionality. To this end, we propose a superpixel based image representation as graph where superpixels are encoded as nodes and nearby superpixels are found using k-nearest neighbor search. This leads to much condense and informative representation of an image over trivial mapping of each pixel to node referred to as dense graph. Further by experimental analysis, we find that the proposed superpixel representation outperforms dense graph representation and results in much more interpretable classifier.

For the solving a Rubik's cube portion, the focus is on combining graph neural network techniques with reinforcement learning techniques. The objective is to see if they provide any benefit over a pure model-free approach which assumes no inductive bias that a Rubik's cube can be thought of as a graph. Here we find the results match but do not exceed the model-free techniques.

\section{Related work}

The widely accepted approach is to represent a graph as a collection of \textit{Vertices, Edges, Adjacency Matrix and Global Graph attributes}, and feed it to a neural network \citep{sanchez2021gentle}. \cite{daigavane2021understanding} details the Graph Convolution layer, which works on message passing among these components of graph with a differentiable  objective. Different proposals have been given for how to allow neural networks to predict properties of graphs at the node level \citep{DBLP:journals/tnn/WuPCLZY21}, edge level, and graph level \citep{DBLP:journals/aiopen/ZhouCHZYLWLS20}.

\subsection{Image Classification using Graph Neural Network}

Our first task is to represent an image as a graph and then test it effectiveness over image classification a graph-level prediction task.The major issue with graph based approach is that it works well on node-level and edge-level tasks but not on graph-level tasks like image classification when the size of the graph is large than few nodes. Several approaches have been proposed for mapping images to graph structures to resolve this issue. Almost all present approaches are ad-hoc and are generally motivated by performance considerations. The most recent notable work is by \cite{DBLP:conf/complexnetworks/NikolentzosTRV20}. The propose King's Graph and Coarsened Graph representations, but each representation has its own shortfalls. Further their analysis is dated since they used MNIST as the dataset for comparison which is shown to be classified well even by learning to discriminate only a single pixel value.

The King's Graph is the dense graph consisting of each pixel as a node and each pixel connected to its 8 neighboring pixel, corresponding to a valid move in the chess board. We test this approach under the name \textbf{Dense graph}. The Coarsened graph is sub-sampled version of King's graph. The authors use community detection algorithms to reduce the number of nodes in a King's graph to a representative subsample of pixels or regions in the Coarsened graph. But their Coarsened graph approach fails miserably even on the MNIST data set and performs much poorly in comparison to King's graph approach. To this end, we propose a new method to encode an image as a graph. We then test the effectiveness of proposed representation in contrast to the Dense graph representation.

\subsection{Reinforcement learning}
The most relevant paper to solving the Rubik's cube using reinforcement learning was \citet{solvingrubikscubes}. In this paper, the authors manage to train the agent to find efficient strategies for solving \(3 \times 3 \times 3\) cubes in a model-free way. \citet{khemani2019solving} proposes using pure graph algorithms with no machine learning to solve the cube effectively. \citet{dqnatari} proposed the deep Q-learning algorithm to great success, and Q-learning seems to be potentially well-suited to environments like the Rubik's cube with a discrete and small action space.

\section{Method}
\subsection{Image Classification}

Our task is to come up with a representation of image as graph. We take the graph-level prediction task of image classification i.e. given an image encoded as graph predict the class of most salient object in the image. We compare 2 approaches for image encoding as graph with a baseline CNN classifier working directly on image matrix.
\begin{enumerate}
    \item Image represented as matrix and classified using a CNN classifier - 2 convolution layers with max pooling and ReLU activation. Classifier consists of 3 fully connected layer with ReLU activation. This is referred to as \textbf{Vanilla CNN (non-Graph representation)}
    \item Image encoded as graph with each pixel mapped to a node, and an edge containing each neighboring node in the graph. We then use Graph convolution layer with node level message passing. This is to referred to as \textbf{Dense Graph (trivial mapping)}
    \item Image encoded as graph with each super pixel as a node. We then use PointNet++ which has 2 phases (1) grouping phase which groups nearby nodes (super-pixels) using k-nearest neighbor search, (2) neighborhood aggregation phase, which uses Graph Neural Network for each node to aggregate information from its direct neighbors. This is referred to as \textbf{SuperPixel Graph (ours)}
\end{enumerate}

\subsubsection{Vanilla CNN (non-Graph representation)}

Vanilla CNN network is given below. Since vanilla CNN network has been extensively researched already, it is expected to give good results. The CNN networks contains 2 parts - one image encoder and classifier. For image encoder we use 2 convolution layers with max pooling and ReLU activation. For Classifier we use 3 fully connected layer with ReLU activation. The reader is referred to appendix~\ref{appendix:vanillacnncode} for details on model definition.

\subsubsection{Dense Graph (trivial mapping)}

In a dense graph, each pixel of the input image is encoded as the a node, and feature vector assigned to each node is RGB value of the pixel. Then all direct neighboring pixels in original image are connected via edges in the encoded graph. This graph has lot of nodes. If the number of nodes is $n$ then number of edges are $O(n)$.

The classifier network uses GraphConv layers, which incorporates the current node features and message passing from all the direct neighbor in an iteration. The difference between Graph Convolution layer (GraphConv) used below and one defined in theory (GNN) is that we add a simple skip-connection to the GNN layer in order to preserve central node information. Hence 3 layer of GraphConv means a node has message passed from all nodes at most distance 3 from it. A Graph Convolution Layer with Skip connection is defined as

\begin{minipage}{\linewidth}
\begin{equation*}
\mathbf{x}_v^{(\ell+1)} = \mathbf{W}^{(\ell + 1)}_1 \mathbf{x}_v^{(\ell)} + \mathbf{W}^{(\ell + 1)}_2 \sum_{w \in \mathcal{N}(v)} \mathbf{x}_w^{(\ell)}
\end{equation*}
\end{minipage}

where $\mathbf{W}^{(\ell + 1)}_{i}$ denotes a trainable weight matrix of shape [$\textit{num\_output\_features}$, $\textit{num\_input\_features}$]. In contrast, a single $Linear$ layer is defined as

\begin{minipage}{\linewidth}
\begin{equation*}
\mathbf{x}_v^{(\ell + 1)} = \mathbf{W}^{(\ell + 1)} \mathbf{x}_v^{(\ell)}
\end{equation*}
\end{minipage}

We then use a readout layer similar to global average pooling which takes all learned node embeddings and returns an embedding for the graph by using the following equation.

\begin{minipage}{\linewidth}
\begin{equation*}
\mathbf{x}_{\mathcal{G}} = \frac{1}{|\mathcal{V}|} \sum_{v \in \mathcal{V}} \mathbf{x}^{(L)}_v
\end{equation*}
\end{minipage}

A linear classifier is finally stacked on top of graph embedding. The reader is referred to appendix~\ref{appendix:densegraphcode} for details on graph convolution layer, readout layer and model definition.

\subsubsection{SuperPixel Graph (ours)}

Superpixels provide a compact representation of image data by grouping perceptually similar pixels together. Hence a localized group of semantically similar pixels is defined a superpixel. This representation is much more informative and condense than corresponding Dense Graph representation. We first find superpixels for an image and encode each superpixel by its centroid pixel location and color. Without explicitly building the edge connections now, we pass the superpixels to  PointNet++ \citep{DBLP:journals/remotesensing/ChenLXPX21} presented in figure~\ref{fig:pointnet}. It has 2 phases - (1) \textbf{grouping phase} which groups nearby nodes (super-pixels) using k-nearest neighbor search. (2) \textbf{neighborhood aggregation phase} which uses Graph Neural Network for each node to aggregate information from its direct neighbors. This is similar to a GraphConv layer but uses a MLP with global max pooling instead of a linear projection to map features from neighboring nodes.

We then use a readout layer similar to global max pooling which takes all learned node embeddings and returns an embedding for the graph. A linear classifier is finally stacked on top of graph embedding. The reader is referred to appendix~\ref{appendix:superpixelcode} for details on PointNet and model definition.

\begin{figure}
     \centering
     \begin{subfigure}[b]{0.26\textwidth}
         \centering
         \includegraphics[width=\textwidth]{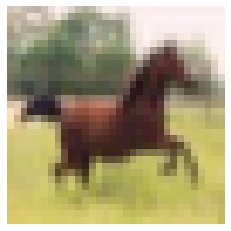}
         \caption{Original image passed to CNN network verbatim\\(non-Graph representation)}
     \end{subfigure}
     \hfill
     \begin{subfigure}[b]{0.26\textwidth}
         \centering
         \includegraphics[width=\textwidth]{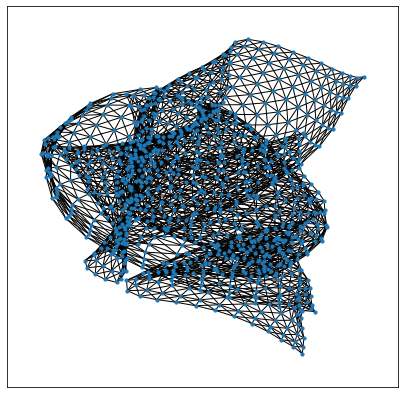}
         \caption{Edge connections in dense graph representation\\(trivial mapping)}
     \end{subfigure}
     \hfill
     \begin{subfigure}[b]{0.26\textwidth}
         \centering
         \includegraphics[width=\textwidth]{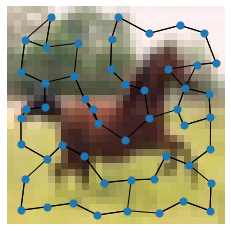}
         \caption{Superpixel as nodes and nearest neighbors as edges\\(ours)}
     \end{subfigure}\\
        \caption{Various Image encoding}
        \label{fig:encode}
\vspace{-0.7cm}
\end{figure}

The final encoded images by the 3 methods presented above are compared in the figure~\ref{fig:encode}. 

\subsection{Reinforcement learning}
In order to more explicitly capture the graph structure of the Rubik's cube, we propose a graph network reinforcement learning (GNRL) algorithm. We call it Node(\(\lambda\)), as it relates to the TD(\(\lambda\)) algorithm \citep{tdlambda}. The pseudocode for the algorithm can be found in Appendix Algorithm~\ref{alg:gnrl}.

This algorithm is specifically designed to be a model-based reinforcement learning algorithm applicable to reversible, goal-based environments with deterministic state transitions and a discrete action space. Here, goal-based means there is no reward needed other than hitting a certain goal state. The environment being deterministic and reversible means that you will always get the same state \(s'\) if taking the same action \(a\) in the same starting state \(s\), and that there is a single action \(a^{R}\) you can take in \(s'\) to get back to \(s\). Reversible environments like this were shown in \citet{qdefects} to have unstable training dynamics using traditional approaches like deep Q-learning, which provides theoretical motivation for our approach.

The basic idea of the algorithm is as follows. There are two main components. (1) We will train \(|A|\) different ``next state'' networks that will predict the next state \(s'\) given an input state \(s\), where \(|A|\) is the size of the action space. The goal of each network is to predict the transition dynamics for a single action. For instance, in the Rubik's cube example, there would be 12 different ``next state'' networks, and one of them would be responsible for predicting the next state given an input state \(s\) and the action being to rotate the front face clockwise. In graph terms, these ``next state'' networks are doing an edge prediction task. Running a given state \(s\) through all \(|A|\) of the action networks predicts all of the available edges out of this node. (2) The second component is the distance network, whose job is to predict how far a given state \(s\) is from the goal. This is similar to a value network \(V(s)\) that is specifically designed for goal-based environments. Only one distance network is needed.

There are several theoretical reasons to believe this kind of architecture could perform well on the Rubik's cube environment. Building the ``next state'' networks should not be too expensive as the state transitions in the cube are deterministic and only involve mapping the input colors to different locations in the output. Assuming the next state networks are trained correctly, this allows us to do planning with essentially arbitrary depth, as given a start state and a long sequence of actions we should know exactly what state we will be in. The distance network as formulated here should also have advantages over a traditional \(Q(s, a)\) or \(V(s)\) formulation. \(Q(s, a)\) and \(V(s)\) are sensitive to the magnitude of the reward signal, whereas this distance function is not, and without a model \(Q(s, a)\) targets are limited to TD(0) error, as they cannot see more than one state ahead.

\section{Experiments}

\subsection{Image Classification}

We use CIFAR-10 dataset with following 10 classes -  \textit{plane, car, bird, cat, deer, dog, frog, horse, ship, truck}. Our train set consists of 90\% of given training data and validation set  consists of remaining training data. We use the entire test data as our test set. In all settings we use Adam optimiser with default parameters for betas and weight\_decay, and $0.001$ learning rate for CNN model and $0.01$ learning rate for graph based models. We use early stopping using validation set loss, with patience of 3 epochs i.e. our model keeps training until its finds 3 epochs where the validation loss increases. It then takes the optimal model based on the least validation loss. We also do a hyperparameter search for k nearest neighbors to choose for $k = 2, 4, 8, 16$. The results are summarized in Table~\ref{tab:imgclass}. The results should be taken with a pinch of salt, since vanilla CNN trains much faster, and was trained for a longer number of epochs than all other models. While dense graph and super-pixel took significantly more time i.e. 50-100x training time. Note this training time is after incorporating speed-up techniques discussed in appendix~\ref{appendix:trainingspeedup}. It is still high compared to CNN since the graph is generated on the fly for a given input image.

\begin{table}
\centering
\begin{tabular}{|c|c|}
\hline Method & Top-1 Accuracy \\
\hline
Random Baseline & 0.1000 \\
Vanilla CNN (non-Graph) & 0.5417 \\
\hline
Dense Graph (trivial mapping) & 0.1943 \\
SuperPixel Graph (ours, k=2) & 0.2107 \\
SuperPixel Graph (ours, k=4) & \textbf{0.2144} \\
SuperPixel Graph (ours, k=8) & 0.2009 \\
SuperPixel Graph (ours, k=16) & 0.1786 \\
\hline
\end{tabular}
\caption{Image Classification accuracy on CIFAR-10 Test, our proposed SuperPixel represenations works better than Dense Graph on all reasonable values of k. The best highlighted in bold}
\label{tab:imgclass}
\vspace{-0.6cm}
\end{table}

Our hypothesis is that SuperPixel network should achieve better results that Dense Graph as a dense graph has lot of redundant information. We expect the vanilla CNN to work the best as this directly operates on the original image space and wherein convolution operation is equivalent to applying a filter over the image. The results show that the Vanilla CNN is best suited for this task and achieves the best accuracy in the least train time. Next, we find that a smaller value of k is better than a larger value and our SuperPixel network using right k value always beats the Dense Graph network. Hence our hypothesis seems to hold well.

Therefore we can conclude that the superpixel based representation with k-nearest neighbors as edges leads to a reasonable graph representation. Note that both graph representations - dense and super-pixel achieve much better results than random baseline and hence demonstrate the effectiveness of learned representation.

\subsubsection{Interpretability of Superpixels}
To further test the effectiveness of Superpixel based representation, we see which superpixels were more salient for the classification. For the saliency method, we use the absolute value of the gradient as the attribution value for each superpixel:

\begin{minipage}{\linewidth}
\begin{equation*}
Attribution_{n_i} = |\frac{\partial F(x)}{\partial x_{n_i}}|
\end{equation*}
\end{minipage}

\begin{figure}[H]
\vspace{-0.2cm}
     \centering
     \begin{subfigure}[b]{0.24\textwidth}
         \centering
         \includegraphics[width=\textwidth]{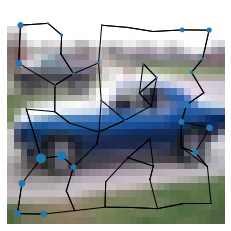}
         \caption{Salience on the tires for car}
     \end{subfigure}
     \hfill
     \begin{subfigure}[b]{0.24\textwidth}
         \centering
         \includegraphics[width=\textwidth]{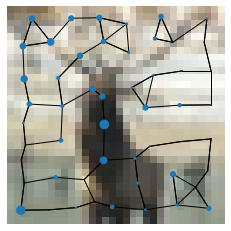}
         \caption{Salience on the torso for deer}
     \end{subfigure}
     \hfill
     \begin{subfigure}[b]{0.24\textwidth}
         \centering
         \includegraphics[width=\textwidth]{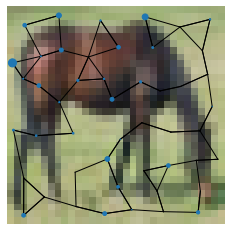}
         \caption{Differentiate foreground from background}
     \end{subfigure}
     \hfill
     \begin{subfigure}[b]{0.24\textwidth}
         \centering
         \includegraphics[width=\textwidth]{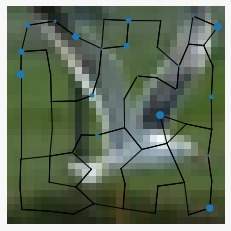}
         \caption{Salience on contrast of wings for bird}
     \end{subfigure}\\
        \caption{Salience of Superpixels denoted by radius of nodes. Note the graph attributes have been calculated using GCN layers via message passing.}
        \label{fig:interpret}
\vspace{-0.6cm}
\end{figure}

Where $x$ is the input and $F(x)$ is the output of the GNN model on input $x$. From figure~\ref{fig:interpret} where radius denotes salience of a superpixel, we can find that the model has learnt well to differentiate foreground from background in all cases, noted by low salience of background superpixels. Further, the model has also learned class specific attributes like tires for car, torso of deer, and contrast of wings of bird for classification.

\subsection{Reinforcement learning}
The hypothesis we wanted to test was that an RL agent that explicitly models graphical structures should outperform a purely model-free agent on an environment with graph properties. The domain chosen to evaluate the new algorithm was that of a custom-made reinforcement learning environment based on a \(2 \times 2 \times 2\) Rubik's cube. The goal here was to gain exposure not just to writing reinforcement learning agents, but also writing the environments from scratch, including choosing appropriate representations for the observation, action, and reward spaces. To do this, we wrote a custom OpenAI gym \citep{openaigym} environment with different options for the state and reward spaces.

To evaluate the final trained agents, each was given 1,000 randomly initialized cubes of varying difficulties.  Stable baselines \citep{stable-baselines3} was used as an off-the-shelf baseline for high quality implementations of various canonical RL algorithms, such as A2C \citep{a2cpaper}, DQN \citep{dqnatari}, and PPO \citep{ppopaper}. For this project, DQN, tabular Q-learning, and the graph network approach were all implemented by hand. Each agent was given between 100,000 and 1,000,000 time steps to train, and this number was treated as a hyperparameter because some agents appeared to perform worse with a higher number of timesteps.

\begin{figure}
\centering
\scalebox{0.57}{
\begin{tikzpicture}
\begin{axis}[
	symbolic x coords={Random, SB A2C, SB DQN, SB PPO, Custom TQN, Custom DQN, Graph Net}, xtick=data, x tick label style={rotate=45}, width=300, ylabel={\text{Cubes solved}}
]
\addplot[ybar, fill=blue]
	coordinates {
	(Random, 87) 
	(SB A2C, 414)
	(SB DQN, 434)
	(SB PPO, 438)
	(Custom TQN, 1000)
	(Custom DQN, 522)
	(Graph Net, 449)
	};
\end{axis}
\end{tikzpicture}
}
\caption{Results of running different trained RL agents on 1,000 randomly initialized cubes}
\label{fig:rlgraph}
\vspace{-0.7cm}
\end{figure}
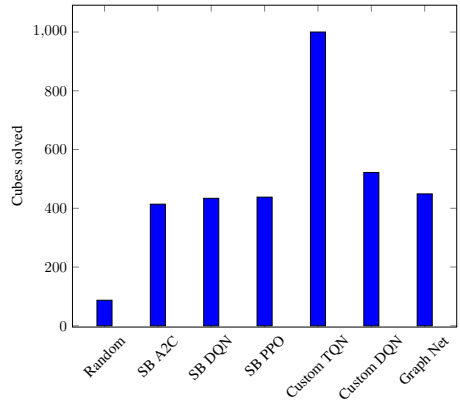

Figure \ref{fig:rlgraph} shows the result of the experiments. Overall, tabular Q-learning performed by far the best out of all agents in terms of sample efficiency, stability of training, and average rewards. The graph network was able to \textit{slightly} outperform the stable baselines \citep{stable-baselines3} implementations, but did not outperform by as much as the custom built DQN and tabular Q-learning agents. Thus, the hypothesis that a model-based graph network should outperform the model-free methods was somewhat nullified, though it did prove to be roughly equal at the task.

Each of the models takes around 2-6 hours to train on a single GPU. Google Colab \citep{googlecolab} was used to develop and train the models. The most impactful idea used in training was to generate rollouts by working backwards from a solved cube, which was inspired by \citet{solvingrubikscubes}. The reason this is so helpful is because solving a Rubik's cube by chance is extremely unlikely, so without training the agent on nearly solved cubes first, almost every rollout will provide only a small reward signal, making training much slower.

\section{Conclusion}

For the first task of representing image as graph, our proposed technique of encoding superpixels as nodes and connecting nearby superpixels by k-nearest neighbor search outperforms the trivial dense graph representation of mapping each pixel to node. Further this representation has smaller dimensionality and leads to much more interpretable results. For the reinforcement learning task, we are able to train agents competitive in performance with the current standard algorithms by using a novel algorithm.

\section{Future Work}
For the image representation task, there is still a large gap between performance of vanilla CNN and graph based image classification model. The major issue with graph based approach is that it works well on node-level and edge-level tasks but not on graph-level tasks like image classification. This may be due to lack of scale invariance in graph. The various layers in a CNN network, can be approximated to looking at an image at various scales, but the same is not true for graph neural network. An approach which adds this scale invariance should provide a large jump in prediction accuracy. Possible future directions involve (1) exploring a hierarchy based collapsing of nodes, where in nodes nearby are collapsed into a single node in a hierarchical manner to achieve scale invariance, (2) exploring graph attention layers.

For the Node\((\lambda)\) algorithm, it would be beneficial to optimize the implementation to work better with larger batches to increase training speed. It may also make sense to consider a tweak to the algorithm that would average over distance estimates for various nodes to provide a less noisy estimate of the shortest path. Finally, we would like to benchmark it on other environments with graph structure.

\bibliographystyle{plainnat}
\bibliography{myrefs.bib}

\newpage
\appendix
\section{Appendix}
\subsection{Training Speed-Up}
\label{appendix:trainingspeedup}
\subsubsection{Mini-batching in Graphs}

Since graphs in graph classification datasets are usually small, we need to batch them to improve GPU utilization.
In the image or language domain, this procedure is typically achieved by rescaling or padding each example into a set of equally-sized shapes which are then grouped.

This strategy, however, is not practicable for GNNs. To accomplish parallelization over a lot of samples, we adopt a different technique. Adjacency matrices are stacked diagonally producing a large network with several isolated subgraphs, and node and target attributes are simply concatenated in the node dimension.

\subsubsection{Scaling Graph Neural Networks}

As the number of nodes and operation on the nodes in a graph increase we have issue of \textbf{neighborhood explosion}.
Cluster-GCN \citep{DBLP:conf/kdd/ChiangLSLBH19} works by first partioning the graph into subgraphs based on graph partitioning algorithms.
With this, GNNs are restricted to solely convolve inside their specific subgraphs, which omits the problem of neighborhood explosion.

However, after the graph is partitioned, some links are removed which may limit the model's performance due to a biased estimation. To address this issue, Cluster-GCN also incorporates between-cluster links inside a mini-batch, which results in the following \textbf{stochastic partitioning scheme}

\begin{figure}[H]
    \centering
    \includegraphics[width=0.7\textwidth]{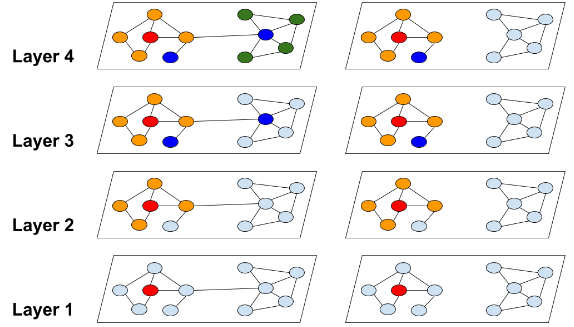}
    \caption{Breaking a larger graph into smaller sub-graph to avoid neighborhood explosion}
    \label{fig:minibatch}
\end{figure}

\begin{figure}[H]
    \centering
    \includegraphics[width=0.7\textwidth]{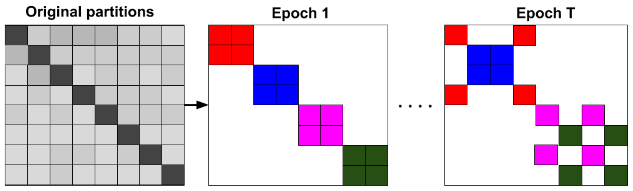}
    \caption{Stochastic partitioning scheme to mitigate effect of lost connections in sub-graphs}
    \label{fig:partition}
\end{figure}

\subsection{Image Classification}

\subsubsection{Vanilla CNN}
\label{appendix:vanillacnncode}
\begin{minted}[breaklines]{python}
class CNN(nn.Module):
    def __init__(self):
        super().__init__()
        self.conv1 = nn.Conv2d(3, 6, 5)
        self.pool = nn.MaxPool2d(2, 2)
        self.conv2 = nn.Conv2d(6, 16, 5)
        self.fc1 = nn.Linear(16 * 5 * 5, 120)
        self.fc2 = nn.Linear(120, 84)
        self.fc3 = nn.Linear(84, 10)

    def forward(self, x):
        x = self.pool(F.relu(self.conv1(x)))
        x = self.pool(F.relu(self.conv2(x)))
        x = torch.flatten(x, 1) # flatten all dimensions except batch
        x = F.relu(self.fc1(x))
        x = F.relu(self.fc2(x))
        x = self.fc3(x)
        return x
\end{minted}

\subsubsection{Dense Graph}

\label{appendix:densegraphcode}

\begin{minted}[breaklines]{python}
class GCN(torch.nn.Module):
    def __init__(self, hidden_channels):
        super(GCN, self).__init__()
        torch.manual_seed(12345)
        self.conv1 = GraphConv(3, hidden_channels)
        self.conv2 = GraphConv(hidden_channels, hidden_channels)
        self.conv3 = GraphConv(hidden_channels, hidden_channels)
        self.lin = Linear(hidden_channels, 10)

    def forward(self, x, edge_index, batch):
        # 1. Obtain node embeddings 
        x = self.conv1(x, edge_index)
        x = x.relu()
        x = self.conv2(x, edge_index)
        x = x.relu()
        x = self.conv3(x, edge_index)

        # 2. Readout layer
        x = global_mean_pool(x, batch)  # [batch_size, hidden_channels]

        # 3. Apply a final classifier
        x = F.dropout(x, p=0.5, training=self.training)
        x = self.lin(x)
        
        return x
\end{minted}

\newpage
\subsubsection{SuperPixel Graph}
\label{appendix:superpixelcode}

\begin{figure}[H]
    \centering
    \includegraphics[width=0.7\textwidth]{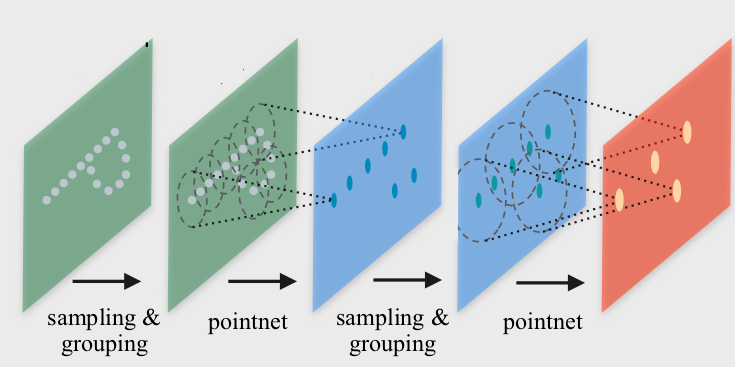}
    \caption{PointNet++ phases}
    \label{fig:pointnet}
\end{figure}

The figure~\ref{fig:pointnet} refers to various phases of PointNet++,

\begin{minted}[breaklines]{python}
class PointNetLayer(MessagePassing):
    def __init__(self, in_channels, out_channels):
        # Message passing with "max" aggregation.
        super().__init__(aggr='max')
        
        # Initialization of the MLP:
        # Here, the number of input features correspond to the hidden node
        # dimensionality plus point dimensionality (=3).
        self.mlp = Sequential(Linear(in_channels + 2, out_channels),
                              ReLU(),
                              Linear(out_channels, out_channels))
        
    def forward(self, h, pos, edge_index):
        # Start propagating messages.
        return self.propagate(edge_index, h=h, pos=pos)
    
    def message(self, h_j, pos_j, pos_i):
        # h_j defines the features of neighboring nodes as shape [num_edges, in_channels]
        # pos_j defines the position of neighboring nodes as shape [num_edges, 3]
        # pos_i defines the position of central nodes as shape [num_edges, 3]

        input = pos_j - pos_i  # Compute spatial relation.

        if h_j is not None:
            # In the first layer, we may not have any hidden node features,
            # so we only combine them in case they are present.
            input = torch.cat([h_j, input], dim=-1)

        return self.mlp(input)  # Apply our final MLP.
        

class PointNet(torch.nn.Module):
    def __init__(self):
        super().__init__()

        torch.manual_seed(12345)
        self.conv1 = PointNetLayer(2, 32)
        self.conv2 = PointNetLayer(32, 32)
        self.classifier = Linear(32, 10)
        
    def forward(self, pos, batch):
        # Compute the kNN graph:
        # Here, we need to pass the batch vector to the function call in order
        # to prevent creating edges between points of different examples.
        # We also add `loop=True` which will add self-loops to the graph in
        # order to preserve central point information.
        edge_index = knn_graph(pos, k=8, batch=batch, loop=True)
        # edge_index = radius_graph(pos, r=12, batch=None, loop=False)
        
        # 3. Start bipartite message passing.
        h = self.conv1(h=pos, pos=pos, edge_index=edge_index)
        h = h.relu()
        h = self.conv2(h=h, pos=pos, edge_index=edge_index)
        h = h.relu()

        # 4. Global Pooling.
        h = global_max_pool(h, batch)  # [num_examples, hidden_channels]
        
        # 5. Classifier.
        return self.classifier(h)

\end{minted}

\newpage
\subsection{Reinforcement Learning}

\begin{algorithm}
\caption{GNRL Node\((\lambda)\) Learning}
\label{alg:gnrl}
\begin{algorithmic}[1]
\State $\textbf{Initialize:}$
\State $A \gets \text{the set of all actions}$
\State $\theta_{D} \gets \text{initialize network to predict a state's distance to goal}$
\For{$a \in A$}
    \State $\theta_{NS}^{a} \gets \text{initialize network to predict next state given action \(a\) taken in input state}$
\EndFor
\State $\lambda \gets \text{hyperparameter used for TD learning}$
\State $b \gets \text{hyperparameter to control how often rollouts are generated by working backwards}$
\State $M \gets \text{maximum number of episodes to train over}$
\State
\Function{DistanceToGoal}{$\lambda, s$}
    \If{$s = G$}
        \State return 0
    \ElsIf{$\lambda = 0$}
        \State $\text{return } \theta_{D}(s)$
    \Else
        \State $s^{min} \gets \text{null}$
        \State $d^{min} \gets \infty$
        \For{$a \in A$}
            \State $s^{pred} \gets \theta_{NS}^{a}(s)$
            \State $d^{pred} \gets \theta_{D}(s^{pred})$
            \If{$d^{pred} < d^{min}$}
                \State $s^{min} \gets s^{pred}$
                \State $d^{min} \gets d^{pred}$
            \EndIf
        \EndFor
        \State $\text{return } 1 + \text{DistanceToGoal}(\lambda-1, s^{min})$
    \EndIf
\EndFunction
\State
\Repeat
    \State $T \gets \text{GenerateRollout(b)}$
    \For{$(s, a, s', \text{is\_done}) \in \text{Reverse}(T)$}
        \State $s^{pred} \gets \theta_{NS}^{a}(s)$
        \State $d^{pred} \gets \theta_{D}(s)$
        \If{$\text{is\_done} = \text{true}$}
            \State $d^{target} \gets 1$
        \Else
            \State $d^{target} \gets 1 + \text{DistanceToGoal}(\lambda, s')$
        \EndIf
        \State $\text{update \(\theta_{NS}^{a}(s)\) using backprop based on \(\mathcal{L}_{NS}(s^{pred}, s')\)}$
        \State $\text{update \(\theta_{D}(s)\) using backprop based on \(\mathcal{L}_{D}(d^{pred}, d^{target})\)}$
    \EndFor
\Until{$iterations = M$}
\end{algorithmic}
\end{algorithm}

\newpage
\subsection{Contributions of each team member}
Overall, this project had an equal contribution from all authors. The image classification sections and cifar-10 Jupyter notebook were primarily written by Naman Goyal, and the reinforcement learning sections and RubiksRL Jupyter notebook were primarily written by David Steiner. The rest of the sections in the writeup were a joint effort.

\end{document}